\documentclass[letterpaper,10pt,conference]{ieeeconf}

\IEEEoverridecommandlockouts
\usepackage{amsmath}
\usepackage{amssymb}
\usepackage{multirow}
\usepackage{booktabs}
\usepackage{adjustbox}
\usepackage{xcolor}
\usepackage{xspace}
\usepackage{siunitx}
\usepackage[hidelinks]{hyperref}
\usepackage{cleveref}
\let\labelindent\relax
\usepackage{enumitem}

\usepackage{caption}
\providecommand{\MethodName}{SpatialDex\xspace}

\crefname{section}{Sec.}{Secs.}
\Crefname{section}{Sec.}{Secs.}
\crefname{subsection}{Sec.}{Secs.}
\Crefname{subsection}{Sec.}{Secs.}

\title{\LARGE \bf 
Reconstruction-Augmented Visuomotor Pre-Training of 3D-Grounded Bimanual Dexterity from RGB-Only Human Videos
}

\author{Xingyi He$^{*}$, Zhiwen Qiu$^{*}$, Adhitya Polavaram$^{*}$, Yunhao Cao$^{*}$, Cory Fan, Julius Samwer,\\
Saleh Kalantari, and Kuan Fang\thanks{All authors are with Cornell University. $^{*}$Equal contribution.}}

\begin{document}
\maketitle
\thispagestyle{empty}
\pagestyle{empty}

\begin{abstract}
We present \MethodName, a framework that learns 3D-grounded bimanual dexterity from RGB-only human videos through reconstruction-augmented visuomotor pre-training. The central idea is to learn human behavior together with its spatial context. From each video, we reconstruct the scene geometry and estimate the hand motion, align them at a shared video-level scale using the hand-size prior of 3D hand estimates. We retain the reconstructed geometry as both a policy input and a prediction target. The policy grounds the token grid of a pre-trained vision-language model in this geometry through gated residual fusion, is trained to predict future scene geometry alongside wrist and finger actions, and generates finger motion conditioned on the predicted trajectories of both wrists. Pre-trained on 1M human-video clips and fine-tuned with a small number of robot demonstrations, \MethodName outperforms the state-of-the-art methods by 16.5 points on nine simulated dexterous tasks across two embodiments and by 38.3 points on six real-world bimanual dexterous tasks. Under a matched comparison, pre-training with reconstructed geometry improves success by 14.2 points over pre-training on hand motion alone. Project page and video: \url{https://spatial-dex.github.io/}
\end{abstract}

\section{Introduction}
\label{sec:intro}

Bimanual dexterous manipulation requires coordination between both hands to stabilize, reorient, and transfer objects while adapting to their shapes and changing positions in the surrounding environment. Geometric representations support this coordination by capturing spatial relationships between hands and objects~\cite{idp3,spatialvla,any3dvla,act3d,dpdactor}, while pre-trained vision-language models (VLMs) can provide complementary visual and semantic priors~\cite{pi05,groot,clip}. Learning such coordination therefore demands both spatial grounding and diverse experience of how the two hands act across objects and scene configurations. Robot demonstrations supply the former through depth sensing and executable commands, but collecting them through teleoperation is costly, which limits the range of tasks, objects, and configurations they cover.

\begin{figure}[t]
\centering
\includegraphics[width=\columnwidth]{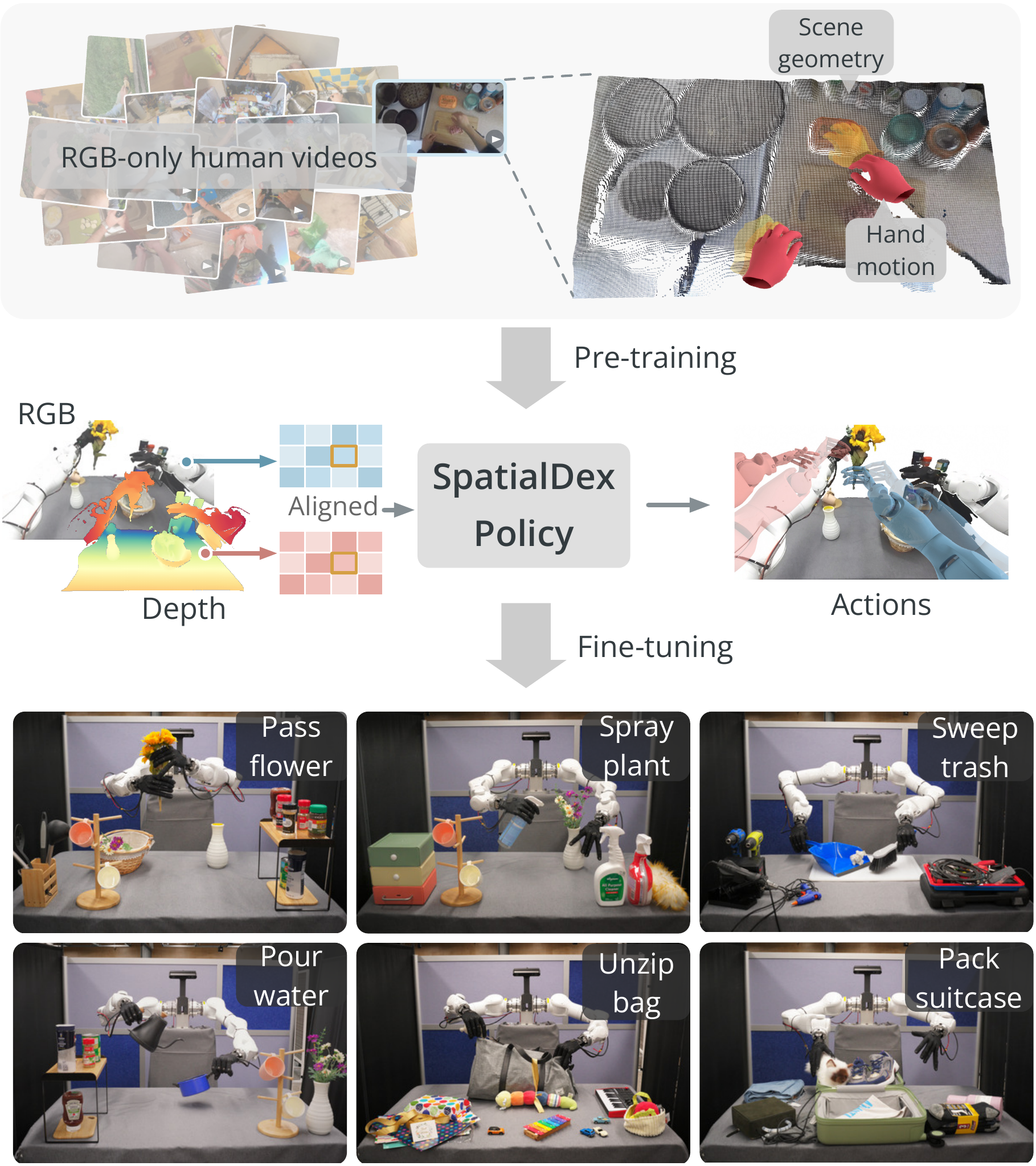}
\caption{
\textbf{\MethodName} reconstructs scene geometry and estimates hand motion from RGB-only human videos and aligns them at a shared video-level scale to jointly supervise policy pre-training. The policy maps RGB-D observations to bimanual dexterous actions and is adapted to specific tasks through fine-tuning on robot demonstrations. The lower panel shows the six bimanual dexterous manipulation tasks for real-world evaluation.}
\label{fig:teaser}
\vspace{-1mm}
\vspace{-0.5\textfloatsep}
\end{figure}

Human videos on the web provide an alternative source of rich everyday activities and physical interactions with diverse environments at scale. However, these RGB-only recordings lack explicit scene geometry or robot-compatible action labels. Learning from these videos therefore calls for recovering the demonstrated hand motion and the scene geometry at a consistent scale and keeping them aligned across time.

Recent methods pre-train vision-language-action (VLA) models on human videos with explicit wrist and finger motion supervision, and transfer learned skills to robots through action-space alignment~\cite{vitra,beingh05,egoscale}. Other approaches exploit 3D cues by predicting point tracks~\cite{arm4r}, adopting shared hand-object keypoint representations~\cite{dpp}, combining calibrated geometry with spatial-language and wrist-motion supervision~\cite{vipavla}, or reconstructing hand-object interactions to synthesize robot trajectories~\cite{deximit}. However, recovering scene geometry at a scale consistent with the recovered hand motion from RGB-only videos remains challenging, as scale errors misalign reconstructed geometry with hand trajectories and compromise spatial supervision. Although some studies integrate geometric and visual features~\cite{any3dvla} or combine spatial and human-motion supervision~\cite{vipavla}, their training objectives provide limited direct supervision for how fine-grained hand behavior relates to changes in the surrounding scene. How to effectively incorporate these cues into human-video pre-training therefore remains underexplored.

In this work, we present \textbf{SpatialDex}, a framework for pre-training 3D-grounded visuomotor policies from large-scale RGB-only human videos. The central idea is to learn human behavior together with its spatial context by reconstructing scene geometry alongside hand motion and using both to supervise policy pre-training. To make reconstructed hand-object distances consistent with the demonstrated motion, we use the hand-size prior in 3D hand estimates to calibrate scene geometry with a shared scale across frames and both hands, as illustrated in Figure~\ref{fig:teaser}. The reconstructed geometry then serves the policy in two roles. As an input, it is combined with the image-patch features of a pre-trained VLM through progressive dual-gated fusion, which adds spatial information while retaining the visual priors of the VLM. As a target, future scene geometry is supervised alongside wrist and finger actions, so that the policy captures spatial relationships and scene dynamics. The policy generates finger motion conditioned on the predicted trajectories of both wrists, and it is fine-tuned to bimanual dexterous tasks with a small number of robot demonstrations.

In summary, the contributions of this work are threefold:
{
\setlength{\leftmargini}{1em}
\begin{itemize}
\item We introduce a data pipeline that aligns reconstructed scene geometry with human hand motion at a shared video-level scale, producing 1M video clips with aligned geometry and motion annotations for policy pre-training.
\item We develop a reconstruction-augmented visuomotor pre-training framework that combines progressive dual-gated fusion of geometric and visual features with future geometry prediction, and generates finger motion conditioned on both wrist trajectories.
\item We evaluate SpatialDex on nine simulated tasks across two embodiments and six real-world tasks, reaching mean success rates of 76.3\% and 81.7\% compared with 59.8\% and 43.3\% for the state-of-the-art methods.
\end{itemize}
}

\section{Related Work}
\label{sec:related}

\noindent\textbf{Learning dexterous behavior from human video.}
Human video provides a scalable source of diverse manipulation experience~\cite{r3m,vip,hrp}, but converting observed behavior into supervision for robot control requires bridging substantial embodiment and action gaps. One line of work recovers human hand motion and uses it to learn transferable action priors~\cite{egomimic}. VITRA~\cite{vitra}, Being-H0.5~\cite{beingh05}, EgoScale~\cite{egoscale}, and UniDex~\cite{unidex} demonstrate that large-scale egocentric or Internet video can provide useful supervision for dexterous policy learning. A complementary direction uses 3D points and correspondences to guide policies or transfer demonstrations across objects~\cite{arm4r,dpp,coil,cordex}. ARM4R~\cite{arm4r} represents human demonstrations through 3D point trajectories, while Dexterous Point Policy~\cite{dpp} uses shared object--hand keypoints to connect human observations with robot actions. These approaches capture either demonstrated hand behavior or the motion of selected scene elements, but the surrounding 3D scene remains only partially represented. Unlike prior approaches that primarily recover action supervision from human videos, SpatialDex additionally reconstructs the surrounding scene and aligns it with the recovered wrist and finger actions.

\noindent\textbf{Scene reconstruction and scale calibration.}
Recent approaches increasingly rely on metric-depth foundation models~\cite{unidepth,depthpro,moge2} to lift RGB human videos into 3D. For example, Dexterous Point Policy~\cite{dpp} uses monocular metric depth to lift object observations into 3D, while VideoManip~\cite{videomanip} uses predicted metric depth as a common spatial reference for hand--object reconstruction. However, predicted metric scale can still deviate from the true physical dimensions, making scene geometry inconsistent with recovered hand motion. To reduce this mismatch, some methods introduce human-hand cues for scale calibration. VIPA-VLA~\cite{vipavla}, for instance, rescales reconstructed geometry using depth ratios between estimated hand joints and corresponding scene points. SpatialDex instead estimates a shared video-level scale from 3D hand correspondences aggregated across frames and both hands, producing scene geometry consistently aligned with recovered wrist and finger actions for large-scale policy pre-training.

\noindent\textbf{Geometry-aware visuomotor policies.}
Explicit geometry has long supported manipulation through point clouds~\cite{idp3,m2t2}, voxels~\cite{peract}, and rendered 3D views~\cite{rvt,rvt2}. While effective for spatial reasoning, these geometry-centric policies typically lack the broad visual and semantic priors of pre-trained vision-language models. Recent VLA methods therefore incorporate depth-derived 3D cues into pre-trained visual representations~\cite{spatialvla,pointvla,geovla,any3dvla}. These methods mainly treat geometry as an additional input to the policy, while the interaction between newly introduced geometric features and pre-trained visual priors is less explicitly modeled. SpatialDex instead explicitly targets this injection problem through reconstruction-augmented pre-training, using progressive geometry fusion and auxiliary forward dynamics prediction to better exploit reconstructed scene geometry.

\begin{figure*}[t]
\centering
\includegraphics[width=\textwidth]{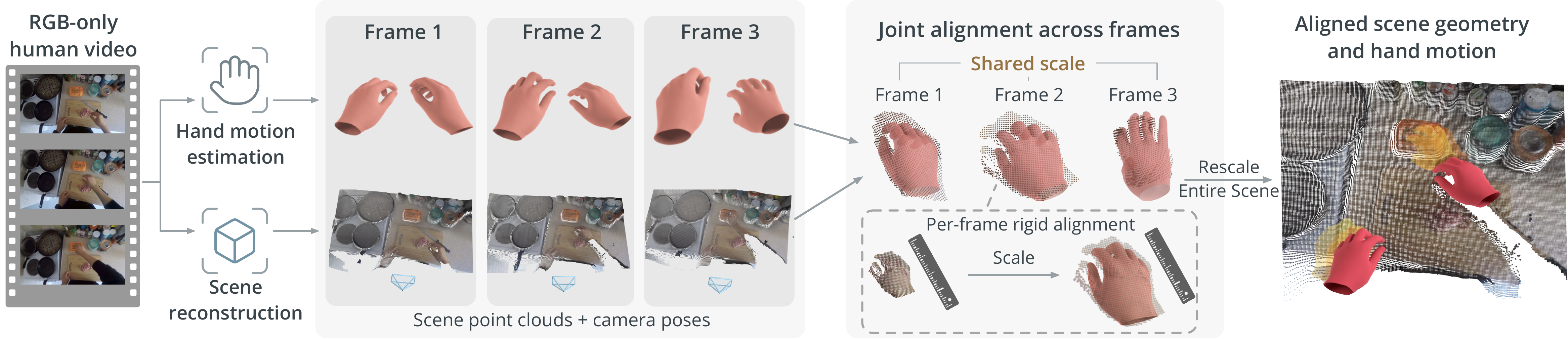}
\caption{
\textbf{Aligned scene and hand reconstruction.} From an RGB-only video, \MethodName recovers hand meshes and wrist trajectories, point clouds and camera poses up to an unknown scale. Aligning the reconstructed hand points with the estimated hand meshes across frames yields one scale shared by the whole video and a rigid transformation per frame, which places the scene geometry and the hand motion in a common frame for pre-training. 
The aligned reconstruction supplies the depth observations, the wrist and finger action targets, and the future geometry targets for policy pre-training.
}
\label{fig:system}
\vspace{-0.5\dbltextfloatsep}
\end{figure*}

\section{Reconstruction-Augmented Visuomotor Pre-Training for Dexterous Manipulation}
\label{sec:model}
\label{sec:method}
\label{sec:arch}

We consider a bimanual robot with two multi-fingered hands that observes an RGB-D image and its proprioceptive state, receives a language instruction, and produces the wrist and finger motion of both hands. Our goal is to pre-train such a policy on human videos so that it learns how the hands move together with the spatial context in which they move, and to transfer it to the robot with a small number of demonstrations. 
In this section, we first describe how the scene and the hands are reconstructed from each RGB-only video and aligned at a shared video-level scale, which turns a video into paired supervision of scene geometry and hand motion (\Cref{sec:data}). 
We then present the 3D-grounded policy, which uses this geometry both as an input fused with the visual features of a pre-trained VLM and as a target for predicting future scene geometry, and generates finger motion conditioned on the predicted trajectories of both wrists (\Cref{sec:policy}). 
Finally, we describe how the policy is fine-tuned on robot demonstrations with sensor depth in the same input format (\Cref{sec:train}).

\subsection{Aligned Scene and Hand Reconstruction}
\label{sec:data}
\label{sec:metric-depth}
\label{sec:pseudo-action}

We construct a large-scale dataset that pairs visual observations with hand motion and scene geometry recovered from RGB-only human videos, as illustrated in Figure~\ref{fig:system}. For each video, we utilize off-the-shelf hand estimators~\cite{hawor} to recover 3D hand meshes and wrist trajectories following VITRA~\cite{vitra}. We additionally estimate scene depth and camera poses using monocular reconstruction methods~\cite{vggtomega}.

Directly combining these estimates, however, introduces two alignment problems. First, although recent monocular reconstruction methods can reliably reconstruct scene geometry and camera motion, the reconstruction is only accurate up to a scaling factor. This makes the reconstructed hand--object distances, wrist trajectories and depth observations inconsistent in scale, resulting in conflicting supervision for learning how hand motion relates to scene geometry. Second, human videos often contain substantial egocentric camera motion, while the robot is deployed with a fixed camera. We address these issues by using the estimated human hands as metric references to calibrate the scene scale, and transforming future geometry and hand motion into the current camera frame to obtain spatially aligned training targets.

\noindent\textbf{Hand-based metric scale calibration.}
To resolve the scale ambiguity, we use the human hand as a metric scale reference. 
Modern MANO-based hand estimators~\cite{hawor} predict a hand mesh with an approximately metric size, while the monocular reconstruction~\cite{vggtomega} recovers scene depth only up to an unknown scale. Given the hand mask rendered from the estimated hand mesh and the camera intrinsics, we back-project the reconstructed depth within the hand region to obtain an estimated hand point cloud. We then estimate the metric scale of the scene reconstruction by aligning these hand points with the predicted MANO hand mesh. While prior works estimate one scaling factor for each frame~\cite{vipavla,deximit}, we jointly solve for a single scale shared across the entire video and a rigid transformation per frame that aligns the estimated mesh with the reconstructed hand points using correspondences from both hands, which aggregates evidence from diverse poses and reduces sensitivity to partial visibility and frame-specific reconstruction errors. After obtaining the scaling factor, we apply the estimated scale to the reconstructed depth maps and camera translations, placing the scene geometry and wrist trajectories at a shared video-level metric scale.

\noindent\textbf{Coordinate frame alignment.}
To address the camera motion misalignment between egocentric videos and static robot cameras, we use the current camera frame as a common spatial reference for scene geometries and hand motions across the prediction horizon. Specifically, during training, we back-project each future depth map into 3D, then transform the point clouds and wrist trajectories from the future camera frame into the current camera frame using known camera poses while keeping finger articulations local for each hand. This removes the effect of camera motion from the training targets and aligns future hand motion and scene geometry with the current observation. Together, our complete alignment pipeline provides spatio-temporally aligned geometric observations, action supervision, and future geometry targets required for bimanual dexterity pre-training.

\subsection{3D-Grounded Visuomotor Policy}
\label{sec:policy}
\label{sec:backproj}
\label{sec:geoenc}
\label{sec:fusion}
\label{sec:future-geo}

\begin{figure*}[t]
\centering
\includegraphics[width=\textwidth]{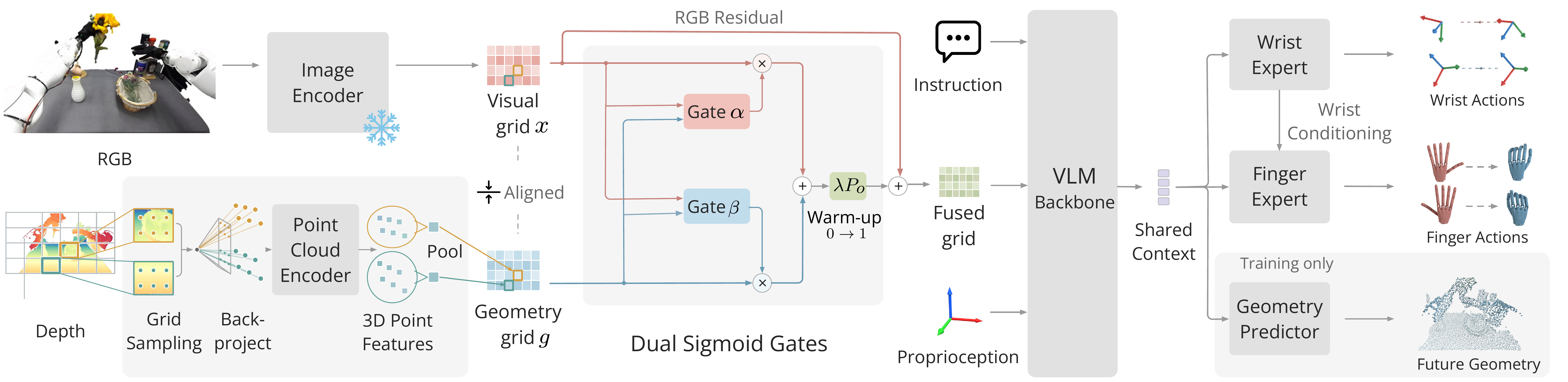}
\caption{
\textbf{3D-grounded visuomotor policy.} Reconstructed geometry enters the policy in two roles. As an input, depth samples from each image patch are encoded jointly in 3D and pooled back to their patches, so that the geometry grid $g$ arrives on the same token grid as the visual features $x$ and is fused into $f$ through two gates and a residual whose weight warms up from zero, leaving the priors of the pre-trained VLM intact. As a target, a training-only predictor estimates future scene geometry from the shared context, which requires the representation to retain spatial structure that the action losses alone do not demand. From the same context, a wrist expert predicts both wrist trajectories and a finger expert predicts finger actions conditioned on them.
}
\label{fig:architecture}
\vspace{-3mm}
\end{figure*}

Using the recovered scene geometry and hand motion from large-scale human videos, we pre-train a geometry-augmented visuomotor policy for bimanual control. As illustrated in Figure~\ref{fig:architecture}, 
given RGB-D observation $o_t = (o_t^{\text{RGB}}, o_t^{\text{depth}})$, proprioception $\rho_t$ consisting of wrist pose and finger articulation, and a task instruction $l$, the policy $\pi(a_{t:t+H}|o_t, \rho_t, l)$ predicts bimanual manipulation action chunks $a_{t:t+H}$. To enable generalizable semantic grounding, we follow prior works~\cite{pi05,spatialvla} to initialize $\pi$ with a pre-trained VLM and structurally expand its modules to enable dexterous manipulation. The key challenge is to effectively incorporate reconstructed geometry without disrupting the VLM's visual priors, enabling the policy to better reason about the 3D scene for action prediction.

\noindent\textbf{Grid-aligned geometric representation.}
While the pre-trained VLM provides strong visual and semantic priors for grounding language instructions and interpreting the observed scene, its image-based representation does not explicitly capture the scene's 3D geometry. The reconstructed depth provides complementary spatial information, but is incompatible with the VLM's image patch features. 
To bridge the two representations without changing the VLM's token structure, we follow grid-aligned approaches~\cite{spatialvla,any3dvla} to encode geometry in 3D and map the resulting features back to their corresponding image patches, as shown in Figure~\ref{fig:architecture}.
Specifically, we sample depth pixels on a regular grid within each image patch and back-project valid samples into the camera frame using the camera intrinsics. We retain the unnormalized 3D coordinates to preserve metric scale and camera-relative position. Since interacting hands and objects may span multiple image patches, we jointly encode valid points across all patches using Sonata~\cite{sonata}. The resulting point features are then mean-pooled according to their original patch assignments, producing a geometric feature grid $g$ that is aligned with the VLM's visual features $x$ and linearly projected to the same dimension. Invalid samples are assigned zero features before pooling. Patches containing no valid samples therefore yield zero geometric features, while their RGB features remain available to the fusion module.

\noindent\textbf{Progressive dual-gated fusion.}
After depth is encoded into patch-level features aligned with the RGB features, a natural approach is to fuse the two modalities through element-wise addition. However, such direct fusion can be undesirable since the pre-trained VLM expects its visual tokens to retain the semantic representation learned during large-scale pre-training. 
We expect the relative importance of visual and geometric features to vary across spatial locations and stages of manipulation. Following prior work on gated residual injection~\cite{flamingo,llama_adapter,any3dvla}, we introduce a dual-gated residual fusion module that independently modulates the visual and geometric features before adding their combination to the original visual representation. A linear warm-up schedule gradually increases the weight of the gated residual update during training, as illustrated in Figure~\ref{fig:architecture}.
Specifically, let $x$ and $g$ denote the visual and geometric features respectively,
then the fused representation is
\begin{equation}
f=x+\lambda P_o\!\left(\alpha\odot x+\beta\odot g\right),
\label{eq:dual-stream-fusion}
\end{equation}
where $P_o$ projects the fused update back to the image-feature dimension, $\odot$ denotes element-wise multiplication, and $\lambda$ controls a warm-up schedule. Modulation weights $\alpha$ and $\beta$ are computed with
\[
\begin{aligned}
\alpha &= \sigma\!\left(F_\alpha([x;g])\right),\\
\beta &= \sigma\!\left(F_\beta([x;g])\right),
\end{aligned}
\]
where $F_\alpha$ and $F_\beta$ are learned MLPs, $\sigma$ is the sigmoid function, and $[x;g]$ denotes channel-wise concatenation. The residual formulation preserves the original visual feature $x$ while allowing appearance and geometry to jointly refine it.
At the beginning of pre-training, the untrained fusion branch can interfere with the pre-trained visual representation. We therefore linearly increase $\lambda$ from zero to one during warm-up and keep it fixed at one thereafter. 
Whereas Any3D-VLA~\cite{any3dvla} learns its 3D representation from synthetic VLA data and visual-grounding data, our geometric encoder and fusion module are learned from large-scale RGB-only human videos covering diverse real-world interactions. The resulting fused feature grid, together with the language instruction and the proprioception $\rho_t$, forms the shared VLM context $c$ for action and forward dynamics prediction.

\noindent\textbf{Forward dynamics prediction.}
Although the policy has access to reconstructed geometry through the fused representation, the strong RGB pathway still provides a shortcut for action prediction, allowing the policy to make limited use of geometric cues. We therefore introduce forward dynamics prediction as an auxiliary objective to explicitly encourage the policy to exploit the reconstructed geometry. Following predictive representation learning~\cite{fvp,dreamvla,pri4r}, a geometry predictor predicts a set of 3D points $\hat{P}_{t+\delta}^{i} = F_{\mathrm{geom}}(c, i, \delta)$ for each image patch $i$ and temporal offset $\delta$, where $c$ is the shared context feature generated by the VLM backbone.

We construct the targets $P_{t+\delta}^i$ using the same grid sampling and back-projection procedure as for the input geometry, ensuring a consistent geometric representation. 
Predicted and target point sets are compared using a geometric distance over valid points:
\begin{equation}
\mathcal{L}_{\mathrm{forward}}
= \frac{1}{H \cdot N_{\text{patch}}}
\sum_{\delta\in[1\dots H]} \sum_{i \in [0\dots N_{\text{patch}} - 1]}
d\left(\widehat{P}_{t+\delta}^i,P_{t+\delta}^i \right),
\label{eq:future-geometry-loss}
\end{equation}
where $H$ is the prediction horizon, $N_{\text{patch}}$ denotes the number of patches in a single frame, and $d(\cdot,\cdot)$ is the symmetric Chamfer distance \begin{equation}
d(A,B)
=
\frac{1}{|A|}
\sum_{\mathbf{a}\in A}
\min_{\mathbf{b}\in B}
\|\mathbf{a}-\mathbf{b}\|_2^2
+
\frac{1}{|B|}
\sum_{\mathbf{b}\in B}
\min_{\mathbf{a}\in A}
\|\mathbf{b}-\mathbf{a}\|_2^2 .
\label{eq:chamfer-distance}
\end{equation}

\noindent\textbf{Wrist-conditioned bimanual action learning.}
\noindent
The geometry-augmented VLM backbone provides a shared representation of appearance, geometry, and language, but dexterous manipulation is inherently asymmetric across different components of the action space. Wrist motion can be recovered relatively reliably from RGB-only videos and transfers naturally across embodiments, whereas fine-grained finger articulation is frequently affected by occlusion and can differ substantially between human hands and the target robot embodiment. Furthermore, finger actions directly determine physical contact and therefore require precise adaptation beyond the motion prior learned from human-video pre-training alone. This suggests a hierarchical decomposition in which the wrist motion provides guidance for refining finger actions. 
Motivated by this insight, we use a wrist expert to jointly predict both wrist trajectories and a finger expert to predict the finger articulation of both hands conditioned on the shared wrist plan, as illustrated in Figure~\ref{fig:architecture}. Let $A^{\text{wrist}} = a_{t:t+H}^{\text{wrist}}$ and $A^{\text{finger}} = a^{\text{finger}}_{t:t+H}$ denote the wrist and finger action chunks, respectively, each containing the actions of both hands.
We factor the action distribution as
\begin{equation}
\resizebox{0.9\columnwidth}{!}{$\displaystyle
p(A^{\text{wrist}},A^{\text{finger}}\mid c)=
p_{\text{wrist}}(A^{\text{wrist}}\mid c)\cdot p_{\text{finger}}(A^{\text{finger}}\mid A^{\text{wrist}},c),
$}
\label{eq:hierarchical-action}
\end{equation}
where $c$ is the shared context feature generated by the VLM backbone. Both experts are implemented as diffusion transformers~\cite{dit,diffusion_policy}. The wrist expert first generates the trajectories of the two wrists,
and then the finger expert predicts the finger actions conditioned on the wrist plan and the shared context $c$. The outputs of both experts are trained jointly using recovered wrist and finger actions for both pre-training and fine-tuning.

\subsection{Human-to-Robot Adaptation}
\label{sec:train}

While human wrist and finger trajectories provide large-scale supervision for coordinated behavior priors, adapting the pre-trained policy to a robot remains challenging due to mismatches in kinematics, appearance, and geometry between the human body and the target embodiment. We therefore fine-tune on a few robot demonstrations, which provide embodiment-specific action and sensor depth supervision, while the geometric encoder, fusion module, and forward dynamics objective are retained unchanged.
Both the pre-training and fine-tuning stages optimize the denoising objectives of the wrist and finger experts together with the forward dynamics objective:
\begin{equation}
\mathcal L=\mathcal L_{\mathrm{wrist}}
+\lambda_{\mathrm{finger}}\mathcal L_{\mathrm{finger}}
+\lambda_{\mathrm{forward}}\mathcal L_{\mathrm{forward}},
\label{eq:total-loss}
\end{equation}
where $\lambda_{\mathrm{finger}}$ and $\lambda_{\mathrm{forward}}$ control the relative weights of finger prediction and forward dynamics prediction. 

During robot fine-tuning, sensor depth is processed by the same geometric encoder, and future RGB-D observations provide supervision for forward dynamics prediction. Embodiment-specific projection and normalization layers convert the action targets to the robot's arm and hand command spaces, with unavailable dimensions masked during training. 

At deployment, the policy takes the current RGB-D observation, camera calibration, language instruction, and robot state and predicts a bimanual action chunk. 
Scene reconstruction and geometry prediction are required only during training, and the forward dynamics decoder is removed at deployment.

\section{Experiments}
\label{sec:exp}

We conduct simulation and real-world experiments to answer three questions. \textbf{Q1:} Does \MethodName outperform existing policies on bimanual dexterous manipulation tasks in simulation and the real world? \textbf{Q2:} Does reconstructed geometry account for the gain, under a matched comparison with motion-only pre-training on the same videos? \textbf{Q3:} How much does each design choice contribute to the performance?

\begin{figure}[t]
\centering
\includegraphics[width=\columnwidth]{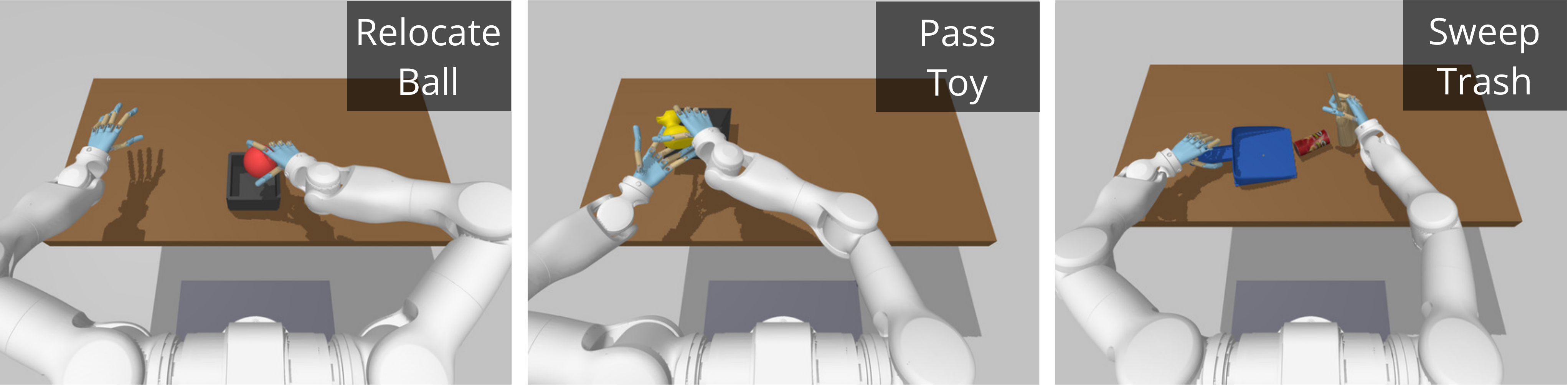}
\caption{\textbf{Wuji--Tianji simulation tasks.} Three Wuji--Tianji tasks are used in addition to the six DexJoCo~\cite{dexjoco} tasks for evaluation in simulation.
}
\label{fig:sim-visualization}
\end{figure}

\subsection{Experimental Setup}
\label{sec:exp-setup}

\noindent\textbf{Pre-training data and setup.}
We initialize the VLM backbone of SpatialDex with pre-trained PaliGemma~\cite{paligemma} weights. We pre-train SpatialDex on 1M human-video clips from Something-Something v.2~\cite{ssv2}, EPIC-Kitchens-100~\cite{epickitchens}, Ego4D~\cite{ego4d}, and Ego-Exo4D~\cite{egoexo4d}, processed as described in Section~\ref{sec:data}. We reconstruct scene geometry using VGGT-Omega~\cite{vggtomega} and estimate hand motion using HaWoR~\cite{hawor} with the MANO~\cite{mano} hand model. Pre-training takes approximately 72 hours on 32 NVIDIA A100 GPUs with a total batch size of 256.
We linearly warm up fusion over 20,000 training steps and set the future geometry prediction offset to $\delta=16$ steps. We predict $H=16$ actions with 5 denoising steps and execute the first 8 at 5\,Hz before replanning.

\noindent\textbf{Baseline methods.}
We compare \MethodName against policies trained from scratch, pre-trained visuomotor models, geometry-aware VLAs, and human-video-pre-trained policies detailed in \Cref{tab:sim-results,tab:real-results}. 
Pre-trained methods are initialized from their public checkpoints and fine-tuned using the same robot demonstrations as \MethodName, with embodiment-specific action heads reinitialized to match the target wrist and finger action spaces.

\begin{table*}[t]
\centering
\caption{\textbf{Simulation success rates (\%).} Avg. columns report unweighted means over the four single-arm Allegro tasks, the two bimanual Allegro tasks, the three bimanual Wuji tasks, and all nine tasks. Values are means over three seeds ($50$ episodes/seed). Best results are shown in \textbf{bold}.}
\label{tab:sim-results}
\setlength{\tabcolsep}{3pt}
\begin{tabular*}{\textwidth}{@{\extracolsep{\fill}}l*{13}{S[table-format=2.1,mode=text,detect-weight=true]}@{}}
\toprule
\multirow{2}{*}{Method} & \multicolumn{6}{c}{Allegro $+$ Franka} & \multicolumn{3}{c}{Wuji $+$ Tianji} & \multicolumn{4}{c}{Avg.} \\
\cmidrule(lr){2-7}\cmidrule(lr){8-10}\cmidrule(lr){11-14}
& \multicolumn{1}{c}{\shortstack{Fold\\Glasses}} & \multicolumn{1}{c}{\shortstack{Hammer\\Nail}} & \multicolumn{1}{c}{\shortstack{Water\\Plant}} & \multicolumn{1}{c}{\shortstack{Pick\\Bucket}} & \multicolumn{1}{c}{\shortstack{Bimanual\\Assembly}} & \multicolumn{1}{c}{\shortstack{Bimanual\\Microwave}} & \multicolumn{1}{c}{\shortstack{Relocate\\Ball}} & \multicolumn{1}{c}{\shortstack{Pass\\Toy}} & \multicolumn{1}{c}{\shortstack{Sweep\\Trash}} & \multicolumn{1}{c}{\shortstack{Unimanual\\Allegro}} & \multicolumn{1}{c}{\shortstack{Bimanual\\Allegro}} & \multicolumn{1}{c}{\shortstack{Bimanual\\Wuji}} & \multicolumn{1}{c}{\shortstack{All\\Tasks}} \\
\midrule
ACT~\cite{aloha}             & 47.3 & 50.0 & 47.3 & 64.0 & 0.0 & 66.0 & 36.7 & 34.0 & 24.7 & 52.2 & 33.0 & 31.8 & 41.1 \\
iDP3~\cite{idp3}             & 10.0 & 18.0 & 48.7 & 54.0 & 0.0 & 38.0 & 20.0 & 14.7 & 10.7 & 32.7 & 19.0 & 15.1 & 23.8 \\
\midrule
$\pi_{0.5}$~\cite{pi05}      & 72.0 & 84.7 & 88.7 & \bfseries 84.0 & 5.3 & 70.0 & 64.7 & 36.0 & 32.7 & 82.4 & 37.7 & 44.5 & 59.8 \\
GR00T N1.5~\cite{grootn15}   & 27.3 & 67.3 & 72.7 & 72.0 & 0.7 & 50.7 & 38.0 & 28.7 & 24.0 & 59.8 & 25.7 & 30.2 & 42.4 \\
ManiFlow~\cite{maniflow}     & 12.7 & 20.0 & 37.3 & 56.7 & 0.0 & 55.3 & 30.7 & 24.0 & 19.3 & 31.7 & 27.7 & 24.7 & 28.4 \\
\midrule
SpatialVLA~\cite{spatialvla} & 8.7 & 14.0 & 24.7 & 26.0 & 0.0 & 18.7 & 18.7 & 11.3 & 8.0 & 18.4 & 9.4 & 12.7 & 14.5 \\
GeoVLA~\cite{geovla}         & 12.0 & 18.7 & 26.7 & 28.0 & 0.7 & 22.0 & 24.7 & 15.3 & 12.0 & 21.4 & 11.4 & 17.3 & 17.8 \\
Any3D-VLA~\cite{any3dvla}    & 2.7 & 25.3 & 46.7 & 30.0 & 0.0 & 43.3 & 36.0 & 26.7 & 20.7 & 26.2 & 21.7 & 27.8 & 25.7 \\
\midrule
VITRA~\cite{vitra}           & 17.3 & 46.7 & 60.0 & 53.3 & 0.0 & 36.7 & 52.0 & 34.7 & 28.0 & 44.3 & 18.4 & 38.2 & 36.5 \\
Being-H0.5~\cite{beingh05}   & 26.7 & 45.3 & 50.0 & 76.7 & 0.0 & 49.3 & 34.7 & 25.3 & 20.0 & 49.7 & 24.7 & 26.7 & 36.4 \\
UniDex~\cite{unidex}         & 10.7 & 16.0 & 28.0 & 27.3 & 1.3 & 20.7 & 22.0 & 16.7 & 10.7 & 20.5 & 11.0 & 16.5 & 17.0 \\
\midrule
\textbf{SpatialDex (Ours)}   & \bfseries 77.3 & \bfseries 94.0 & \bfseries 91.3 & 79.3 & \bfseries 16.7 & \bfseries 84.0 & \bfseries 88.7 & \bfseries 75.3 & \bfseries 80.0 & \bfseries 85.5 & \bfseries 50.4 & \bfseries 81.3 & \bfseries 76.3 \\
\bottomrule
\end{tabular*}
\vspace{-0.5\dbltextfloatsep}
\end{table*}

\noindent\textbf{Evaluation environments and protocols.}
All methods use the same evaluation initializations. A trial is successful if the task is completed within the allowed execution time.

\noindent\textit{Simulation.}
We use 100 training demonstrations per task.
We evaluate six DexJoCo~\cite{dexjoco} tasks detailed in \Cref{tab:sim-results} using Allegro hands with Franka arms. We also introduce three tasks: Relocate Ball, Pass Toy, and Sweep Trash using Wuji hands with Tianji arms to evaluate the performance of the policy on an embodiment with more degrees of freedom (Figure~\ref{fig:sim-visualization}). The first four DexJoCo tasks are unimanual, whereas Bimanual Assembly and Bimanual Microwave require bimanual coordination. We evaluate each method and ablation variant over three random seeds, with 50 episodes per task and seed.

\noindent\textit{Real-world.}
For each task, we collect 50 demonstration episodes through teleoperation. We evaluate SpatialDex and seven baselines on six tasks using a bimanual Wuji--Tianji platform (Table~\ref{tab:real-results}; the real-world Sweep Trash is a separate task from the simulated one), with ten trials per method and task.

\setcounter{topnumber}{1}
\setcounter{dbltopnumber}{1}

\begin{figure*}[t]
\centering
\includegraphics[width=\textwidth]{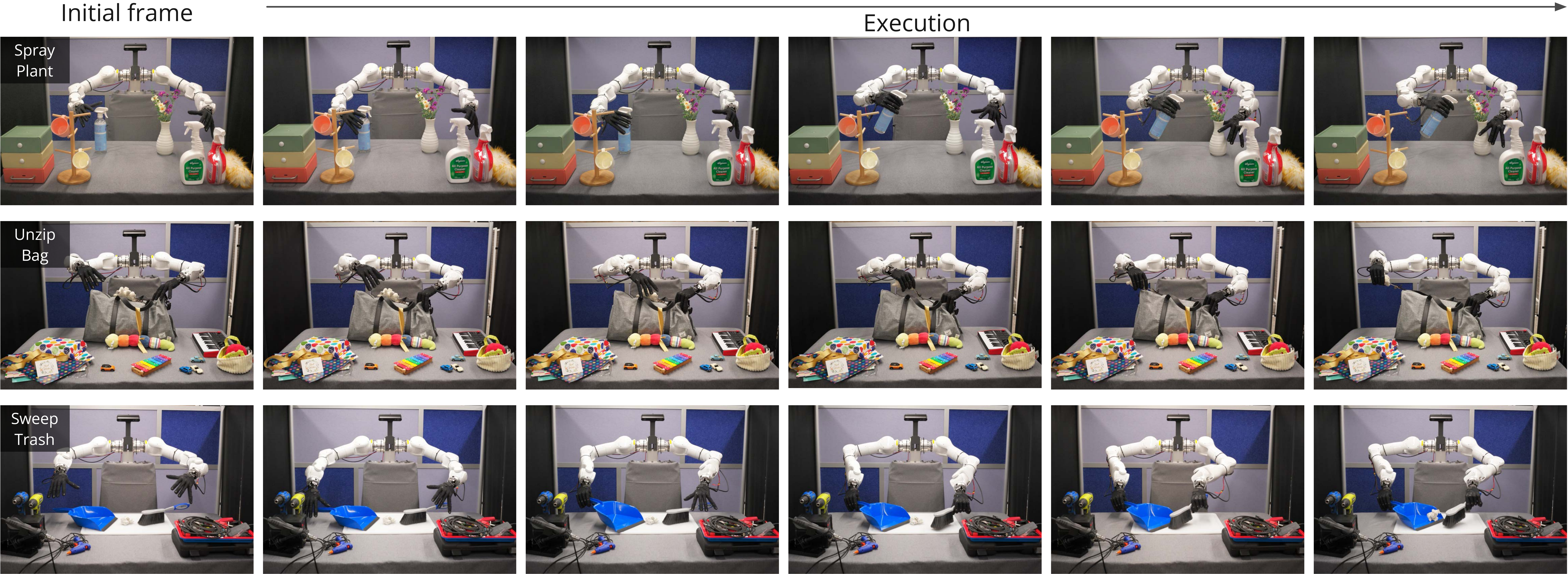}
\caption{
\textbf{Real-world rollouts.} \MethodName on Spray Plant, Unzip Bag, and Sweep Trash with Wuji hands and Tianji arms, which involve tool actuation, a deformable object, and coordinated two-hand sweeping. Each row runs from the initial frame on the left to task completion on the right.
}
\label{fig:real-world}
\end{figure*}

\subsection{Comparative Results}
\label{sec:exp-sim}
\label{sec:exp-real}

\noindent\textbf{Simulation.}
As shown in \Cref{tab:sim-results}, \MethodName achieves 76.3\% mean success while the strongest baseline $\pi_{0.5}$ achieves 59.8\%. 
The largest gains occur on tasks requiring bimanual coordination, Bimanual Assembly, Bimanual Microwave, Relocate Ball, Pass Toy, and Sweep Trash, while improvements on unimanual tasks are modest and $\pi_{0.5}$ performs better on the Pick Bucket task (84.0\% versus 79.3\%). Neither human-video pre-training alone (VITRA) nor explicit geometric input alone (iDP3) is sufficient here, which motivates the controlled studies below that isolate each component. Bimanual Assembly remains the most challenging task for all methods, but \MethodName achieves a higher success rate than the strongest baseline (16.7\% versus 5.3\%).

\begin{table}[t]
\centering
\caption{\textbf{Real-world success counts.} Successful trials out of 10. Best results are shown in \textbf{bold}.}
\label{tab:real-results}
\begin{adjustbox}{max width=\columnwidth,center}
\setlength{\tabcolsep}{2.5pt}
\newcommand{\realcount}[2]{\makebox[1em][r]{#1}/\makebox[1em][l]{#2}}
\begin{tabular}{lccccccc}
\toprule
Method & \shortstack{Pass\\Flower} & \shortstack{Pour\\Water} & \shortstack{Spray\\Plant} & \shortstack{Unzip\\Bag} & \shortstack{Sweep\\Trash} & \shortstack{Pack\\Suitcase} & Total \\
\midrule
ACT~\cite{aloha}            & \realcount{4}{10} & \realcount{4}{10} & \realcount{1}{10} & \realcount{4}{10} & \realcount{4}{10} & \realcount{5}{10} & \realcount{22}{60} \\
iDP3~\cite{idp3}              & \realcount{1}{10} & \realcount{1}{10} & \realcount{2}{10} & \realcount{1}{10} & \realcount{0}{10} & \realcount{3}{10} & \realcount{8}{60} \\
\midrule
$\pi_{0.5}$~\cite{pi05}     & \realcount{4}{10} & \realcount{4}{10} & \realcount{5}{10} & \realcount{5}{10} & \realcount{2}{10} & \realcount{6}{10} & \realcount{26}{60} \\
\midrule
GeoVLA~\cite{geovla}        & \realcount{2}{10} & \realcount{2}{10} & \realcount{4}{10} & \realcount{2}{10} & \realcount{3}{10} & \realcount{4}{10} & \realcount{17}{60} \\
Any3D-VLA~\cite{any3dvla} & \realcount{3}{10} & \realcount{4}{10} & \realcount{2}{10} & \realcount{2}{10} & \realcount{3}{10} & \realcount{4}{10} & \realcount{18}{60} \\
\midrule
VITRA~\cite{vitra}          & \realcount{3}{10} & \realcount{2}{10} & \realcount{5}{10} & \realcount{4}{10} & \realcount{3}{10} & \realcount{5}{10} & \realcount{22}{60} \\
UniDex~\cite{unidex}        & \realcount{3}{10} & \realcount{4}{10} & \realcount{2}{10} & \realcount{4}{10} & \realcount{1}{10} & \realcount{5}{10} & \realcount{19}{60} \\
\midrule
\textbf{SpatialDex (Ours)}     & \textbf{\realcount{8}{10}} & \textbf{\realcount{8}{10}} & \textbf{\realcount{9}{10}} & \textbf{\realcount{8}{10}} & \textbf{\realcount{7}{10}} & \textbf{\realcount{9}{10}} & \textbf{\realcount{49}{60}} \\
\bottomrule
\end{tabular}
\end{adjustbox}
\vspace{-0.5\textfloatsep}
\end{table}

\noindent\textbf{Real-world.}
As shown in \Cref{tab:real-results}, \MethodName achieves a total of 49 successes in 60 trials (81.7\%), outperforming the strongest baseline $\pi_{0.5}$, which achieves 26/60 (43.3\%). In particular, Spray Plant and Unzip Bag reach 9/10 and 8/10 successes, respectively, compared with 5/10 for the strongest competing methods on each task. These results show that the gains extend to physical tool actuation and deformable-object manipulation using sensor depth at deployment. Sweep Trash remains the most challenging real-world task for SpatialDex at 7/10, while still exceeding the best baseline at 4/10. We show representative rollouts of \MethodName in \Cref{fig:real-world}.

\subsection{Pre-Training and Ablation Studies}
\label{sec:exp-ablation}
\begin{table}[t]
\centering
\caption{\textbf{Pre-training and component ablations.} Success rates (\%). Means over three seeds (50 episodes per task and seed).}
\label{tab:ablation}
\begin{adjustbox}{max width=\columnwidth,center}
\footnotesize
\setlength{\tabcolsep}{1.2pt}
\begin{tabular}{@{}l*{7}{S[table-format=2.1,mode=text,detect-weight=true]}@{}}
\toprule
\multirow{2}{*}{Variant} & \multicolumn{3}{c}{Allegro $+$ Franka} & \multicolumn{3}{c}{Wuji $+$ Tianji} & \multicolumn{1}{c}{Avg.} \\
\cmidrule(lr){2-4}\cmidrule(lr){5-7}\cmidrule(lr){8-8}
& \multicolumn{1}{c}{\shortstack{Hammer\\Nail}} & \multicolumn{1}{c}{\shortstack{Bimanual\\Assembly}} & \multicolumn{1}{c}{\shortstack{Bimanual\\Microwave}} & \multicolumn{1}{c}{\shortstack{Relocate\\Ball}} & \multicolumn{1}{c}{\shortstack{Pass\\Toy}} & \multicolumn{1}{c}{\shortstack{Sweep\\Trash}} \\
\midrule
\textbf{SpatialDex (Ours)} & \bfseries 94.0 & \bfseries 16.7 & \bfseries 84.0 & \bfseries 88.7 & \bfseries 75.3 & \bfseries 80.0  & \bfseries 73.1 \\
\midrule
No human pre-training & 62.7 & 4.0 & 52.0 & 64.0 & 44.7 & 46.0  & 45.6 \\
Motion-only pre-training & 80.0 & 8.7 & 66.7 & 77.3 & 58.0 & 62.7  & 58.9 \\
Per-frame alignment & 90.7 & 8.0 & 68.0 & 87.0 & 55.3 & 74.0  & 63.8 \\
\midrule
RGB-only input & 50.0 & 2.3 & 41.3 & 55.3 & 40.0 & 32.0  & 36.8 \\
2D depth encoding & 72.0 & 6.7 & 60.0 & 74.0 & 50.7 & 56.7  & 53.3 \\
Concat RGB + geometry & 82.0 & 6.0 & 69.3 & 73.3 & 60.7 & 62.7  & 59.0 \\
No fusion warm-up & 87.3 & 15.0 & 83.0 & 88.7 & 67.0 & 68.0  & 68.2 \\
No forward dynamics loss & 87.3 & 12.0 & 76.0 & 82.7 & 65.3 & 67.3  & 65.1 \\
\midrule
Joint wrist-finger expert & 90.0 & 12.7 & 78.7 & 85.3 & 66.7 & 73.3  & 67.8 \\
No wrist-trajectory input & 92.0 & 13.3 & 80.7 & 86.0 & 68.0 & 76.0  & 69.3 \\
\bottomrule
\end{tabular}
\end{adjustbox}
\vspace{-0.5\textfloatsep}
\end{table}

Table~\ref{tab:ablation} evaluates all Wuji--Tianji tasks and the three DexJoCo tasks: a unimanual nail-hammering task, a bimanual peg-insertion task, and a bimanual task involving door operation, food placement, and button pressing. These three tasks also show the largest gains over the strongest per-task baselines in Table~\ref{tab:sim-results}.
All variants follow the protocol as stated in Section~\ref{sec:exp-setup}.

\noindent\textbf{Reconstructed geometry during pre-training.}
We first isolate the contribution of reconstructed scene geometry during human-video pre-training. Motion-only pre-training uses the same RGB videos, wrist and finger targets, and pre-training budget as SpatialDex, but removes geometric inputs and forward dynamics targets. The variant without human pre-training directly fine-tunes the pre-trained VLM on robot demonstrations. During robot fine-tuning, all variants use the same RGB-D architecture, demonstrations, objectives, and optimization schedule, with matched initialization for newly introduced geometry modules.

Motion-only pre-training improves mean success from 45.6\% to 58.9\% over the variant without human pre-training and exceeds $\pi_{0.5}$ at 48.9\% on the same six tasks, which supports the benefit of human-motion supervision. Full SpatialDex further improves mean success to 73.1\%, a gain of 14.2 percentage points over motion-only pre-training. This isolates the benefit of reconstructed geometry as both input and supervision during pre-training.

\noindent\textbf{Joint alignment across frames.}
Replacing the shared video-level alignment with independent per-frame alignment reduces mean success from 73.1\% to 63.8\%, a drop of 9.3 percentage points. The largest reductions occur on Pass Toy and Bimanual Microwave, both of which require coordinated motion over multiple stages. These results support constructing scene-action supervision with a shared video-level scale, which keeps reconstructed geometry and wrist trajectories metrically consistent across frames.

\noindent\textbf{Geometric representation and supervision.}
We next study how reconstructed geometry is represented, fused, and supervised while keeping the data and optimization budgets fixed. Removing geometric inputs and fusion while retaining forward dynamics supervision reduces mean success to 36.8\%, showing that forward dynamics supervision alone cannot compensate for the absence of geometric observations. Encoding depth directly as a 2D channel or replacing dual-gated fusion with feature concatenation also reduces success on every task. These comparisons support explicit 3D encoding and selective fusion over simply supplying depth to the policy. Removing fusion warm-up reduces mean success from 73.1\% to 68.2\%, a drop of 4.9 percentage points, with the largest losses on Sweep Trash (sim) and Pass Toy. This supports progressively introducing the geometric update into the pre-trained visual representation. Removing only the forward dynamics objective reduces mean success by 8.0 percentage points while leaving geometric inputs unchanged, isolating the benefit of auxiliary supervision on how the scene evolves.

\noindent\textbf{Coordinated action prediction.}
Replacing the separate wrist and finger experts with a single joint wrist--finger expert reduces mean success by 5.3 percentage points. Keeping separate wrist and finger experts but removing wrist-trajectory conditioning reduces mean success by 3.8 percentage points, with observations and current-state inputs unchanged. Both variants incur their largest losses on Pass Toy, where finger closure must be coordinated with wrist motion that brings the hands and object together. These results support conditioning finger actions on the predicted wrist trajectories, which provide intended future motion beyond the shared current observation.

\section{Conclusion}
\label{sec:conclusion}

SpatialDex converts RGB-only human videos into aligned scene geometry and hand-motion supervision for bimanual visuomotor pre-training. It incorporates reconstructed geometry as both a representation of the current scene and supervision for forward dynamics prediction. The resulting policy achieves higher mean success than the compared baselines across the evaluated simulation and real-world tasks. Controlled comparisons further show the benefit of incorporating reconstructed geometry beyond motion-only pre-training. These results support jointly using scene geometry and hand motion to transfer human manipulation experience to bimanual robot control.

\noindent\textbf{Limitations.}
The quality of our pre-training supervision depends on the accuracy of hand estimation and scene reconstruction. Therefore, the shared-scale alignment cannot fully correct errors in estimated hand size, pose, or camera motion. At deployment, the policy requires depth observations which can limit its applicability when reliable depth is unavailable or degraded by sensing noise. Adapting the policy to a new robot also requires embodiment-specific demonstrations to account for differences in kinematics, action spaces, and control interfaces.

\bibliographystyle{IEEEtran}
\bibliography{references}

@article{paligemma,
  author  = {Beyer, Lucas and Steiner, Andreas and Pinto, Andr{\'e} Susano and others},
  title   = {{PaliGemma}: A versatile {3B} {VLM} for transfer},
  journal = {arXiv preprint arXiv:2407.07726},
  year    = {2024}
}

@inproceedings{pi05,
  author    = {{Physical Intelligence} and Black, Kevin and Brown, Noah and Darpinian, James and Dhabalia, Karan and Driess, Danny and Esmail, Adnan and Equi, Michael and Finn, Chelsea and Fusai, Niccolo and Galliker, Manuel Y. and Ghosh, Dibya and Groom, Lachy and Hausman, Karol and Ichter, Brian and Jakubczak, Szymon and Jones, Tim and Ke, Liyiming and LeBlanc, Devin and Levine, Sergey and Li-Bell, Adrian and Mothukuri, Mohith and Nair, Suraj and Pertsch, Karl and Ren, Allen Z. and Shi, Lucy Xiaoyang and Smith, Laura and Springenberg, Jost Tobias and Stachowicz, Kyle and Tanner, James and Vuong, Quan and Walke, Homer and Walling, Anna and Wang, Haohuan and Yu, Lili and Zhilinsky, Ury},
  title     = {{$\pi_{0.5}$}: A {VLA} with Open-World Generalization},
  booktitle = {{CoRL}},
  year      = {2025}
}

@inproceedings{vitra,
  author    = {Li, Qixiu and Deng, Yu and Liang, Yaobo and Luo, Lin and Zhou, Lei and Yao, Chengtang and Zeng, Lingqi and Feng, Zhiyuan and Liang, Huizhi and Xu, Sicheng and Zhang, Yizhong and Chen, Xi and Chen, Hao and Sun, Lily and Chen, Dong and Yang, Jiaolong and Guo, Baining},
  title     = {Scalable Vision-Language-Action Model Pretraining for Robotic Dexterous Manipulation with Real-Life Human Activity Videos},
  booktitle = {{ICRA}},
  year      = {2026}
}

@inproceedings{aloha,
  author    = {Tony Z. Zhao and Vikash Kumar and Sergey Levine and Chelsea Finn},
  title     = {Learning Fine-Grained Bimanual Manipulation with Low-Cost Hardware},
  booktitle = {{RSS}},
  year      = {2023}
}

@article{diffusion_policy,
  author  = {Chi, Cheng and Xu, Zhenjia and Feng, Siyuan and Cousineau, Eric and Du, Yilun and Burchfiel, Benjamin and Tedrake, Russ and Song, Shuran},
  title   = {Diffusion Policy: Visuomotor Policy Learning via Action Diffusion},
  journal = {{IJRR}},
  year    = {2024}
}

@inproceedings{egomimic,
  author    = {Kareer, Simar and Patel, Dhruv and Punamiya, Ryan and Mathur, Pranay and Cheng, Shuo and Wang, Chen and Hoffman, Judy and Xu, Danfei},
  title     = {{EgoMimic}: Scaling Imitation Learning via Egocentric Video},
  booktitle = {{ICRA}},
  year      = {2025}
}

@article{coil,
  author  = {Cao, Yunhao and Bhaumik, Zubin and Jia, Jessie and He, Xingyi and Fang, Kuan},
  title   = {Correspondence-Oriented Imitation Learning: Flexible Visuomotor Control with {3D} Conditioning},
  journal = {arXiv preprint arXiv:2512.05953},
  year    = {2025}
}

@inproceedings{cordex,
  author    = {He, Xingyi and Polavaram, Adhitya and Cao, Yunhao and Deshmukh, Om and Wang, Tianrui and Zhou, Xiaowei and Fang, Kuan},
  title     = {Generate, Transfer, Adapt: Learning Functional Dexterous Grasping from a Single Human Demonstration},
  booktitle = {{ICRA}},
  pages     = {22388--22395},
  year      = {2026},
  doi       = {10.1109/ICRA57385.2026.11697272}
}

@inproceedings{r3m,
  author    = {Nair, Suraj and Rajeswaran, Aravind and Kumar, Vikash and Finn, Chelsea and Gupta, Abhinav},
  title     = {{R3M}: A Universal Visual Representation for Robot Manipulation},
  booktitle = {{CoRL}},
  year      = {2022}
}

@inproceedings{vip,
  author    = {Ma, Yecheng Jason and Sodhani, Shagun and Jayaraman, Dinesh and Bastani, Osbert and Kumar, Vikash and Zhang, Amy},
  title     = {{VIP}: Towards Universal Visual Reward and Representation via Value-Implicit Pre-Training},
  booktitle = {{ICLR}},
  year      = {2023}
}

@inproceedings{hrp,
  author    = {Srirama, Mohan Kumar and Dasari, Sudeep and Bahl, Shikhar and Gupta, Abhinav},
  title     = {{HRP}: Human Affordances for Robotic Pre-Training},
  booktitle = {{RSS}},
  year      = {2024}
}

@article{groot,
  author  = {{NVIDIA}},
  title   = {{GR00T N1}: An Open Foundation Model for Generalist Humanoid Robots},
  journal = {arXiv preprint arXiv:2503.14734},
  year    = {2025}
}

@misc{grootn15,
  author       = {{NVIDIA}},
  title        = {{Isaac GR00T N1.5}: An Open Foundation Model for Generalized Humanoid Robot Reasoning and Skills},
  howpublished = {NVIDIA model release},
  year         = {2025}
}

@article{beingh05,
  author  = {Luo, Hao and Wang, Ye and Zhang, Wanpeng and Zheng, Sipeng and Xi, Ziheng and Xu, Chaoyi and Xu, Haiweng and Yuan, Haoqi and Zhang, Chi and Wang, Yiqing and Feng, Yicheng and Lu, Zongqing},
  title   = {{Being-H0.5}: Scaling Human-Centric Robot Learning for Cross-Embodiment Generalization},
  journal = {arXiv preprint arXiv:2601.12993},
  year    = {2026}
}

@article{maniflow,
  author  = {Yan, Ge and Zhu, Jiyue and Deng, Yuquan and Yang, Shiqi and Qiu, Ri-Zhao and Cheng, Xuxin and Memmel, Marius and Krishna, Ranjay and Goyal, Ankit and Wang, Xiaolong and Fox, Dieter},
  title   = {{ManiFlow}: A General Robot Manipulation Policy via Consistency Flow Training},
  journal = {arXiv preprint arXiv:2509.01819},
  year    = {2025}
}

@article{unidex,
  author  = {Zhang, Gu and Xu, Qicheng and Zhang, Haozhe and Ma, Jianhan and He, Long and Bao, Yiming and Ping, Zeyu and Yuan, Zhecheng and Lu, Chenhao and Yuan, Chengbo and Liang, Tianhai and Tian, Xiaoyu and Shao, Maanping and Zhang, Feihong and Ding, Mingyu and Gao, Yang and Zhao, Hao and Zhao, Hang and Xu, Huazhe},
  title   = {{UniDex}: A Robot Foundation Suite for Universal Dexterous Hand Control from Egocentric Human Videos},
  journal = {arXiv preprint arXiv:2603.22264},
  year    = {2026}
}

@article{idp3,
  title   = {Generalizable Humanoid Manipulation with 3D Diffusion Policies},
  author  = {Yanjie Ze and Zixuan Chen and Wenhao Wang and Tianyi Chen and Xialin He and Ying Yuan and Xue Bin Peng and Jiajun Wu},
  year    = {2024},
  journal = {arXiv preprint arXiv:2410.10803}
}

@article{dpdactor,
  author  = {Tsung-Wei Ke and Nikolaos Gkanatsios and Katerina Fragkiadaki},
  title   = {3D Diffuser Actor: Policy Diffusion with 3D Scene Representations},
  journal = {arXiv preprint arXiv:2402.10885},
  year    = {2024}
}

@inproceedings{peract,
  author    = {Mohit Shridhar and Lucas Manuelli and Dieter Fox},
  title     = {Perceiver-Actor: A Multi-Task Transformer for Robotic Manipulation},
  booktitle = {{CoRL}},
  year      = {2022}
}

@inproceedings{rvt,
  author    = {Goyal, Ankit and Xu, Jie and Guo, Yijie and Blukis, Valts and Chao, Yu-Wei and Fox, Dieter},
  title     = {{RVT}: Robotic View Transformer for 3D Object Manipulation},
  booktitle = {{CoRL}},
  year      = {2023}
}

@article{pointvla,
  author  = {Li, Chengmeng and Wen, Junjie and Peng, Yan and Peng, Yaxin and Feng, Feifei and Zhu, Yichen},
  title   = {{PointVLA}: Injecting the 3D World into Vision-Language-Action Models},
  journal = {arXiv preprint arXiv:2503.07511},
  year    = {2025}
}

@article{geovla,
  author  = {Sun, Lin and Xie, Bin and Liu, Yingfei and Shi, Hao and Wang, Tiancai and Cao, Jiale},
  title   = {{GeoVLA}: Empowering 3D Representations in Vision-Language-Action Models},
  journal = {arXiv preprint arXiv:2508.09071},
  year    = {2025}
}

@article{spatialvla,
  author  = {Delin Qu and Haoming Song and Qizhi Chen and Yuanqi Yao and Xinyi Ye and Yan Ding and Zhigang Wang and JiaYuan Gu and Bin Zhao and Dong Wang and Xuelong Li},
  title   = {{SpatialVLA}: Exploring Spatial Representations for Visual-Language-Action Model},
  journal = {arXiv preprint arXiv:2501.15830},
  year    = {2025}
}

@inproceedings{act3d,
  author    = {Theophile Gervet and Zhou Xian and Nikolaos Gkanatsios and Katerina Fragkiadaki},
  title     = {{Act3D}: 3D Feature Field Transformers for Multi-Task Robotic Manipulation},
  booktitle = {{CoRL}},
  year      = {2023}
}

@inproceedings{m2t2,
  author    = {Yuan, Wentao and Murali, Adithyavairavan and Mousavian, Arsalan and Fox, Dieter},
  title     = {{M2T2}: Multi-Task Masked Transformer for Object-centric Pick and Place},
  booktitle = {{CoRL}},
  year      = {2023}
}

@inproceedings{vggtomega,
  author    = {Wang, Jianyuan and Chen, Minghao and Zhang, Shangzhan and Karaev, Nikita and Sch{\"o}nberger, Johannes and Labatut, Patrick and Bojanowski, Piotr and Novotny, David and Vedaldi, Andrea and Rupprecht, Christian},
  title     = {{VGGT-$\Omega$}},
  booktitle = {{CVPR}},
  year      = {2026},
  pages     = {21486--21499}
}

@article{mano,
  author  = {Javier Romero and Dimitrios Tzionas and Michael J. Black},
  title   = {Embodied Hands: Modeling and Capturing Hands and Bodies Together},
  journal = {{TOG}},
  year    = {2017}
}

@inproceedings{ssv2,
  author    = {Goyal, Raghav and Kahou, Samira Ebrahimi and Michalski, Vincent and Materzy{\'{n}}ska, Joanna and Westphal, Susanne and Kim, Heuna and Haenel, Valentin and Fruend, Ingo and Yianilos, Peter and Mueller-Freitag, Moritz and Hoppe, Florian and Thurau, Christian and Bax, Ingo and Memisevic, Roland},
  title     = {The ``Something Something'' Video Database for Learning and Evaluating Visual Common Sense},
  booktitle = {{ICCV}},
  year      = {2017}
}

@article{epickitchens,
  author  = {Damen, Dima and Doughty, Hazel and Farinella, Giovanni Maria and Furnari, Antonino and Kazakos, Evangelos and Ma, Jian and Moltisanti, Davide and Munro, Jonathan and Perrett, Toby and Price, Will and Wray, Michael},
  title   = {Rescaling Egocentric Vision: Collection, Pipeline and Challenges for {EPIC-KITCHENS-100}},
  journal = {{IJCV}},
  year    = {2022}
}

@inproceedings{ego4d,
  author    = {Grauman, Kristen and Westbury, Andrew and Byrne, Eugene and Chavis, Zachary and Furnari, Antonino and Girdhar, Rohit and Hamburger, Jackson and Jiang, Hao and Liu, Miao and Liu, Xingyu and Martin, Miguel and Nagarajan, Tushar and Radosavovic, Ilija and Ramakrishnan, Santhosh Kumar and Ryan, Fiona and Sharma, Jayant and Wray, Michael and Xu, Mengmeng and Xu, Eric Zhongcong and Zhao, Chen and Bansal, Siddhant and Batra, Dhruv and Cartillier, Vincent and Crane, Sean and Do, Tien and Doulaty, Morrie and Erapalli, Akshay and Feichtenhofer, Christoph and Fragomeni, Adriano and Fu, Qichen and Gebreselasie, Abrham and Gonz{\'{a}}lez, Cristina and Hillis, James and Huang, Xuhua and Huang, Yifei and Jia, Wenqi and Khoo, Weslie and Kol{\'{a}}{\v{r}}, J{\'{a}}chym and Kottur, Satwik and Kumar, Anurag and Landini, Federico and Li, Chao and Li, Yanghao and Li, Zhenqiang and Mangalam, Karttikeya and Modhugu, Raghava and Munro, Jonathan and Murrell, Tullie and Nishiyasu, Takumi and Price, Will and Ruiz, Paola and Ramazanova, Merey and Sari, Leda and Somasundaram, Kiran and Southerland, Audrey and Sugano, Yusuke and Tao, Ruijie and Vo, Minh and Wang, Yuchen and Wu, Xindi and Yagi, Takuma and Zhao, Ziwei and Zhu, Yunyi and Arbel{\'{a}}ez, Pablo and Crandall, David and Damen, Dima and Farinella, Giovanni Maria and Fuegen, Christian and Ghanem, Bernard and Ithapu, Vamsi Krishna and Jawahar, C. V. and Joo, Hanbyul and Kitani, Kris and Li, Haizhou and Newcombe, Richard and Oliva, Aude and Park, Hyun Soo and Rehg, James M. and Sato, Yoichi and Shi, Jianbo and Shou, Mike Zheng and Torralba, Antonio and Torresani, Lorenzo and Yan, Mingfei and Malik, Jitendra},
  title     = {{Ego4D}: Around the World in 3{,}000 Hours of Egocentric Video},
  booktitle = {{CVPR}},
  year      = {2022}
}

@inproceedings{egoexo4d,
  author    = {Kristen Grauman and others},
  title     = {{Ego-Exo4D}: Understanding Skilled Human Activity from First- and Third-Person Perspectives},
  booktitle = {{CVPR}},
  year      = {2024}
}

@article{dexjoco,
  title={DexJoCo: A Benchmark and Toolkit for Task-Oriented Dexterous Manipulation on MuJoCo},
  author={Wang, Hanwen and Zhao, Weizhi and Wang, Xiangyu and Huang, Siyuan and Lin, He and Zheng, Boyuan and Xu, Rongtao and Wang, Gang and Mu, Yao and Wang, He and Fan, Lue and Li, Hongsheng and Zhang, Zhaoxiang and Tan, Tieniu},
  journal={arXiv preprint arXiv:2605.16257},
  year={2026}
}

@article{arm4r,
  author  = {Niu, Dantong and Sharma, Yuvan and Xue, Haoru and Biamby, Giscard and Zhang, Junyi and Ji, Ziteng and Darrell, Trevor and Herzig, Roei},
  title   = {{Pre-training Auto-regressive Robotic Models with 4D Representations}},
  journal = {arXiv preprint arXiv:2502.13142},
  year    = {2025}
}

@article{vipavla,
  author  = {Feng, Yicheng and Zhang, Wanpeng and Wang, Ye and Luo, Hao and Yuan, Haoqi and Zheng, Sipeng and Lu, Zongqing},
  title   = {{Spatial-Aware VLA Pretraining through Visual-Physical Alignment from Human Videos}},
  journal = {arXiv preprint arXiv:2512.13080},
  year    = {2025}
}

@article{any3dvla,
  author  = {Fan, Xianzhe and Deng, Shengliang and Wu, Xiaoyang and Lu, Yuxiang and Li, Zhuoling and Yan, Mi and Zhang, Yujia and Zhang, Zhizheng and Wang, He and Zhao, Hengshuang},
  title   = {{Any3D-VLA: Enhancing VLA Robustness via Diverse Point Clouds}},
  journal = {arXiv preprint arXiv:2602.00807},
  year    = {2026}
}

@article{egoscale,
  author  = {Zheng, Ruijie and Niu, Dantong and Xie, Yuqi and Wang, Jing and Xu, Mengda and Jiang, Yunfan and Casta{\~{n}}eda, Fernando and Hu, Fengyuan and Tan, You Liang and Fu, Letian and Darrell, Trevor and Huang, Furong and Zhu, Yuke and Xu, Danfei and Fan, Linxi},
  title   = {{EgoScale: Scaling Dexterous Manipulation with Diverse Egocentric Human Data}},
  journal = {arXiv preprint arXiv:2602.16710},
  year    = {2026}
}

@article{deximit,
  author  = {Mu, Juncheng and Yang, Sizhe and Bao, Yiming and Bae, Hojin and Wei, Tianming and Xu, Linning and Li, Boyi and Xu, Huazhe and Pang, Jiangmiao},
  title   = {{DexImit: Learning Bimanual Dexterous Manipulation from Monocular Human Videos}},
  journal = {arXiv preprint arXiv:2602.10105},
  year    = {2026}
}

@article{dpp,
  author  = {Kim, Beomjun and Park, Seong Hyeon and Sim, Seunghoon and Moon, Seungjun and Lee, Sanghyeok and Shin, Jinwoo},
  title   = {{Dexterous Point Policy: Learning Point-based Dexterous Hand Policies from Human Demonstrations}},
  journal = {arXiv preprint arXiv:2606.10614},
  year    = {2026}
}

@article{hawor,
  author  = {Zhang, Jinglei and Deng, Jiankang and Ma, Chao and Potamias, Rolandos Alexandros},
  title   = {{HaWoR: World-Space Hand Motion Reconstruction from Egocentric Videos}},
  journal = {arXiv preprint arXiv:2501.02973},
  year    = {2025}
}

@article{dreamvla,
  author  = {Zhang, Wenyao and Liu, Hongsi and Qi, Zekun and Wang, Yunnan and Yu, Xinqiang and Zhang, Jiazhao and Dong, Runpei and He, Jiawei and Lu, Fan and Wang, He and Zhang, Zhizheng and Yi, Li and Zeng, Wenjun and Jin, Xin},
  title   = {{DreamVLA: A Vision-Language-Action Model Dreamed with Comprehensive World Knowledge}},
  journal = {arXiv preprint arXiv:2507.04447},
  year    = {2025}
}

@article{pri4r,
  author  = {Kim, Jisoo and Cho, Jungbin and Chu, Sanghyeok and Bal, Ananya and Kim, Jinhyung and Lee, Gunhee and Lee, Sihaeng and Kim, Seung Hwan and Han, Bohyung and Lee, Hyunmin and Jeni, Laszlo A. and Kim, Seungryong},
  title   = {{Pri4R: Learning World Dynamics for Vision-Language-Action Models with Privileged 4D Representation}},
  journal = {arXiv preprint arXiv:2603.01549},
  year    = {2026}
}

@article{fvp,
  author  = {Hou, Chengkai and Ze, Yanjie and Fu, Yankai and Gao, Zeyu and Hu, Songbo and Yu, Yue and Zhang, Shanghang and Xu, Huazhe},
  title   = {{4D Visual Pre-training for Robot Learning}},
  journal = {arXiv preprint arXiv:2508.17230},
  year    = {2025}
}

@article{sonata,
  author  = {Wu, Xiaoyang and DeTone, Daniel and Frost, Duncan and Shen, Tianwei and Xie, Chris and Yang, Nan and Engel, Jakob and Newcombe, Richard and Zhao, Hengshuang and Straub, Julian},
  title   = {{Sonata: Self-Supervised Learning of Reliable Point Representations}},
  journal = {arXiv preprint arXiv:2503.16429},
  year    = {2025}
}

@article{videomanip,
  author  = {Hongyi Chen and Tony Dong and Tiancheng Wu and Liquan Wang and Yash Jangir and Yaru Niu and Yufei Ye and Homanga Bharadhwaj and Zackory Erickson and Jeffrey Ichnowski},
  title   = {Dexterous Manipulation Policies from {RGB} Human Videos via {3D} Hand-Object Trajectory Reconstruction},
  journal = {arXiv preprint arXiv:2602.09013},
  year    = {2026}
}

@inproceedings{unidepth,
  author    = {Luigi Piccinelli and Yung-Hsu Yang and Christos Sakaridis and Mattia Segu and Siyuan Li and Luc Van Gool and Fisher Yu},
  title     = {{UniDepth}: Universal Monocular Metric Depth Estimation},
  booktitle = {{CVPR}},
  year      = {2024}
}

@inproceedings{depthpro,
  author    = {Aleksei Bochkovskii and Ama{\"e}l Delaunoy and Hugo Germain and Marcel Santos and Yichao Zhou and Stephan R. Richter and Vladlen Koltun},
  title     = {{Depth Pro}: Sharp Monocular Metric Depth in Less Than a Second},
  booktitle = {{ICLR}},
  year      = {2025}
}

@article{moge2,
  author  = {Ruicheng Wang and Sicheng Xu and Yue Dong and Yu Deng and Jianfeng Xiang and Zelong Lv and Guangzhong Sun and Xin Tong and Jiaolong Yang},
  title   = {{MoGe-2}: Accurate Monocular Geometry with Metric Scale and Sharp Details},
  journal = {arXiv preprint arXiv:2507.02546},
  year    = {2025}
}

@inproceedings{rvt2,
  author    = {Ankit Goyal and Valts Blukis and Jie Xu and Yijie Guo and Yu-Wei Chao and Dieter Fox},
  title     = {{RVT-2}: Learning Precise Manipulation from Few Demonstrations},
  booktitle = {{RSS}},
  year      = {2024}
}

@inproceedings{clip,
  author    = {Alec Radford and Jong Wook Kim and Chris Hallacy and Aditya Ramesh and Gabriel Goh and Sandhini Agarwal and Girish Sastry and Amanda Askell and Pamela Mishkin and Jack Clark and Gretchen Krueger and Ilya Sutskever},
  title     = {Learning Transferable Visual Models from Natural Language Supervision},
  booktitle = {{ICML}},
  year      = {2021}
}

@inproceedings{dit,
  author    = {William Peebles and Saining Xie},
  title     = {Scalable Diffusion Models with Transformers},
  booktitle = {{ICCV}},
  year      = {2023}
}

@article{flamingo,
  title={Flamingo: a visual language model for few-shot learning},
  author={Alayrac, Jean-Baptiste and Donahue, Jeff and Luc, Pauline and Miech, Antoine and Barr, Iain and Hasson, Yana and Lenc, Karel and Mensch, Arthur and Millican, Katherine and Reynolds, Malcolm and others},
  journal={Advances in neural information processing systems},
  volume={35},
  pages={23716--23736},
  year={2022}
}

@article{llama_adapter,
  title = {LLaMA-Adapter: Efficient Fine-tuning of Language Models with Zero-init Attention},
  author={Zhang, Renrui and Han, Jiaming and Liu, Chris and Gao, Peng and Zhou, Aojun and Hu, Xiangfei and Yan, Shilin and Lu, Pan and Li, Hongsheng and Qiao, Yu},
  journal={arXiv preprint arXiv:2303.16199},
  year={2023}
}

\end{document}